# An Application-Agnostic Automatic Target Recognition System Using Vision Language Models

Anthony Palladino, Dana Gajewski, Abigail Aronica, Patryk Deptula, Alexander Hamme,
Seiyoung C. Lee, Jeff Muri, Todd Nelling, Michael A. Riley, Brian Wong, Margaret Duff

The Charles Stark Draper Laboratory, Inc.

## Abstract

We present a novel Automatic Target Recognition (ATR) system using open-vocabulary object detection and classification models. A primary advantage of this approach is that target classes can be defined just before runtime by a non-technical end user, using either a few natural language text descriptions of the target, or a few image exemplars, or both. Nuances in the desired targets can be expressed in natural language, which is useful for unique targets with little or no training data. We also implemented a novel combination of several techniques to improve performance, such as leveraging the additional information in the sequence of overlapping frames to perform tubelet identification (i.e., sequential bounding box matching), bounding box re-scoring, and tubelet linking. Additionally, we developed a technique to visualize the aggregate output of many overlapping frames as a mosaic of the area scanned during the aerial surveillance or reconnaissance, and a kernel density estimate (or heatmap) of the detected targets. We initially applied this ATR system to the use case of detecting and clearing unexploded ordinance on airfield runways and we are currently extending our research to other real-world applications.

## Introduction

Automatic Target Recognition (ATR) systems leverage advanced algorithms, machine learning (ML), and artificial intelligence (AI) techniques to autonomously recognize objects of interest in real time. The use of deep neural networks (DNNs) in ATR models have improved the accuracy and speed of detection and classification; however, approaches using DNNs require retraining the entire model when adding a new class and this process can take days or weeks, even on high-performance computing clusters (Dhillon and Verma 2020). In this paper, we utilize Open Vocabulary Object Detection (OVOD) as the basis for a novel ATR system that allows non-technical users to define and add new target classes just before inference time, using only language descriptions and/or image exemplars with no retraining. This allows users to flexibly and quickly adapt to evolving threats or environments, which is not possible with current ATR systems.

To demonstrate our ATR system, we applied it to the detection and classification of unexploded ordinance (UXO). In a conflict scenario, there will not be sufficient time or resources to retrain traditional models to detect a new class (i.e., previously unseen UXO or vehicle type) that it was not trained on. Our solution leverages the massive semantic knowledge contained in OVOD models to allow users to describe what they are looking for in words, as they would communicate with their teammates. Further, the ATR system is generalizable, in that the same system can be used for one application (e.g., UXO clearance) one day, and re-purposed by the end user for another application (e.g., vehicle tracking) the next day. This work is a first step in optimizing human-AI integration for object recognition by using language to convey complex information instead of complicated technical features that are not comprehensible to most users.

## Related Work

ATR systems using machine learning are predominantly based on DNN architectures, e.g., sensing and control of crop-harvesting robots (Liu and Liu 2024), identifying underwater objects such as wreckage (Khan, et al. 2024), and synthetic aperture radar (SAR) for remote sensing and guidance (Li, et al. 2024). These systems are often restricted to the fixed vocabulary, or set of object classes, in the training set tailored to their specific application. Likewise, academic advancements in model performance are often evaluated using fixed object sets, such as the 80 classes in the COCO dataset, thus the pre-trained models are often unsuitable for direct application and custom labeled datasets tailored to intended applications are needed.

State-of-the-art vision-language models (VLMs) show a remarkable ability to overcome this fixed target set limitation and can accurately detect a wide range of objects in a zero-shot setting (Zhang, et al. 2024). Only a subset of VLMs can perform open-vocabulary recognition at region (i.e., bounding box) level as opposed to whole-image level.

Current state-of-the-art models include: DITO, which uses contrastive learning without pseudo-labeling (Kim, Angelova and Kuo 2023); OV-DQUO, which mitigates detector confidence bias between base and novel categories (Wang, et al. 2024); and LaMI-DETR, which uses GPT and T5 to establish inter-category relationships (Du, et al. 2024). YOLO-World is a recent advancement that enhances You Only Look Once (YOLO) architectures with open-vocabulary detection capabilities (Cheng, et al. 2024). MM-OVOD, which was the top-performing OVOD model when we started developing our ATR system, allows specifying the desired class with natural language text descriptions (rather than just a class name), or image exemplars, or both (Kaul, Xie and Visserman 2023).

Object detection models, including the OVOD models introduced in the preceding paragraph, are typically evaluated on a frame-by-frame or bounding-box basis, rather than a per-object basis, and often fail to leverage information in neighboring frames when used to analyze video. Some detection models are specifically designed to process video streams, typically implementing feature aggregation from neighboring frames, (Zhu, et al. 2017) and (Wu, et al. 2019), however they are often computationally intensive and slow. To our knowledge, there are no video-specific OVOD models. Instead, for our system, we chose to implement a purely post-processing solution to leverage information in neighboring frames, so that we can keep our system modular, couple it with our OVOD model, and implement newer OVOD models as they become available. Some entirely post-processing solutions include Seq-NMS (Han, et al. 2016) which links predictions across frames and refines their detections, one that focuses on improving detections of fast-moving objects (Sabater, Montesano and Murillo 2020), and Seq-BBox-Matching (Belhassen, et al. 2019), which we slightly modified for our ATR system.

## ATR System Overview

Our system is agnostic to the VLM "under-the-hood", and newer/better models can be swapped in as they become available. This paper demonstrates our approach using a publicly-available pre-trained MM-OVOD model (Kaul, Xie and Visserman 2023). Their model architecture and training recipe used a two-stage CenterNet2 model (Zhou, Koltun and Krahneb 2021) with a ResNet50 backbone (He, et al. 2016) pre-trained on ImageNet21k-P (Ridnik, et al. 2021). It achieved a mask average precision over "rare" classes (APr) of 27.3% and a mask average precision over all classes (mAP) of 33.1% on the LVIS v1.0 open-vocabulary benchmark, which was one of the top-performing OVOD models when we started this work in early 2024.

---

[2] US Army Reserve photo by Lt. Col. Angela Wallace. Source: https://www.usar.army.mil/News/Images/igphoto/2001900807/

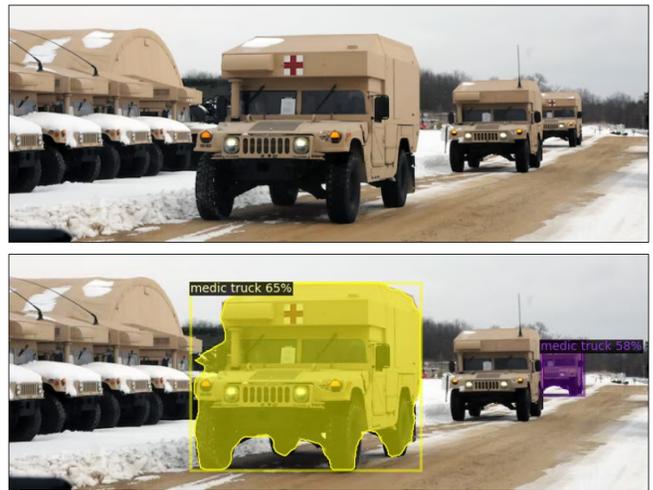

Figure 1: Example of a nuanced class, where the medic truck is only slightly different from the regular truck. *Top*: Original image with two medic trucks interspersed with non-medic trucks. *Bottom*: Demonstration of our system correctly detecting only the two desired medic trucks.

Our system also contains functionality to (*i*) specify the desired target classes, (*ii*) leverage sequential frames, and (*iii*) build a mosaic visualization along with kernel density estimation to improve results on static target applications. Each of these functionalities is discussed in detail below.

Figure 1 shows an example[2] of our system where a user's desired target class is only slightly different from an undesired class. Natural language text descriptions facilitate the specification of the desired nuance. To the best of our knowledge, ours is the first operational–ATR System that leverages recent advancements in OVOD using VLMs, which is ready to be deployed by end users.

## Application to UXO Clearing

UXO clearance involves the identification, localization, and destruction of potentially explosive hazards. The Joint Service Explosive Ordinance Disposal (JEOD) Community, via the Office of Naval Research (ONR), sought commercial ML-enabled solutions for Rapid Large Area Clearance (RLAC) to help automate and accelerate what is normally a very manual and dangerous process (Office of Naval Research 2024). This opportunity is focused on developing algorithms to help mitigate explosive hazards at stand-off distances. Specifically, it is focused on UXO removal from an airstrip using overhead drones flying at an altitude of 8-10 meters, equipped with visible imagery (RGB color) cameras collecting video in a pre-set lawnmower pattern; however, algorithms are expected to generalize to other locations and conditions.

The demonstration stage of the competition process was a week-long test and evaluation (T&E) event at Eglin Air Force Base where offerors applied their algorithms (trained earlier on a previously provided dataset) to real images collected during the event. Each test run was comprised of various types of UXO scattered across an otherwise empty and flat airstrip in unique configurations and each offeror provided their real-time results on the identification and location of each object detected.

**Performance Metrics**

We use three metrics to quantify the performance of our approach, which were defined by ONR for the RLAC competition. The probability of detecting a UXO is defined as:

$$P_d = \frac{N_d^{\text{True}}}{N_T} \quad (1)$$

where $N_d^{\text{True}}$ is the number of detected UXO objects, and $N_T$ is the ground truth number of target UXO opportunities. Here, the classification is not considered, such that any class with $c \in \mathcal{C}_{\text{UXO}}$ is considered a true UXO detection, even if it is misclassified.

The false alarm density for detections is defined as:

$$D_d^{\text{FA}} = \frac{N_d^{\text{False}}}{A} \quad (2)$$

where $N_d^{\text{False}}$ is the number of falsely-detected objects that do not correspond to the ground truth UXO locations and are classified with $c \in \mathcal{C}_{\text{UXO}}$, and $A$ is the area of the airfield runway encompassed by all the image frames in the video. In this study, we used data consisting of runway segments 140 feet (35 m) wide by 400 ft (122 m) long, resulting in an area, $A = 4{,}270$ m².

The probability of correct classification is defined as:

$$P_c = \frac{N_c^{\text{Correct}}}{N_d^{\text{True}}} \quad (3)$$

which measures the proportion of correct classifications given our true detections.

## Defining Classes of Interest

Table 1 lists the classes we used in the ATR system described in this paper. The UXO classes include high-explosive anti-tank detonating submunitions, NATO-standard artillery projectiles that are used in field guns, howitzers, and gun-howitzers, scatterable bombs roughly the size of a baseballs, and high-explosive anti-tank/anti-personnel submunitions with self-destruct-after-impact mechanisms.

MM-OVOD, introduced above, allows users to define classes using natural language text descriptions, or image exemplars, or both. For each class, we produced natural language text descriptions describing the visible attributes of the classes. We took care to include unique *visible* qualities specific to each class, such as a color, geometric shape, and the distinct physical appearance. The optional image exemplars depicting the target classes are composed of images that are depicting two- and three-dimensional representations of objects, and enhanced images with background noise added to represent the realistic imagery photographed by a drone. Table 1 shows the some of the text descriptions and image exemplars for the UXO classes.

Table 1: Example text descriptions and image exemplars used to define our classes for the UXO ATR application.

| Class | Text Description | Image Exemplars |
|---|---|---|
| 155MM | • "155MM is shaped like a bullet and is olive to brown in color with yellow markings"<br>• "155MM is an artillery shell that is dark green, grey, or olive in color" | 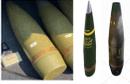 |
| BLU26 | • "BLU26 is a small round object that appears similar to a small rock and is green or blue"<br>• "BLU26 are spherical bombs olive or silver in color" | 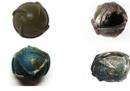 |
| BLU63 | • "BLU63 is a bomb that looks like a baseball and silver in color"<br>• "BLU63 is a submunition with two hemispheres held together by a crimped flange" | 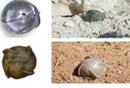 |
| BLU97 | • "BLU97 looks like an elongated blue or yellow beer can with a cloth napkin attached to an end"<br>• "BLU97 is a blue or yellow cylinder with a cone shaped cloth on the end that is either white or olive green in color" | 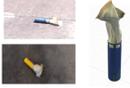 |
| PTAB2.5KO | • "PTAB2.5KO looks like a brown or silver tube with many long metallic fins that protrude outwards"<br>• "PTAB2.5KO is a brown or silver metallic torpedo-shaped object with pointed tails at one end" | 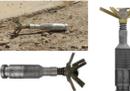 |
| ROCKEYE | • "ROCKEYE has an orange or blue cylindrical-shaped body and a smaller cylindrical nose, with small fins on the white end"<br>• "ROCKEYE is a two-piece object with an orange solid tube-shaped piece and a white tail with three very small fins" | 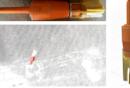 |

## Leveraging Sequential Frames

We implement two complementary approaches: (*i*) sequential bounding box matching, adapted from Belhassen et al. (2019) and (*ii*) mosaic with kernel density estimation (our novel approach). We use these two approaches in concert. Existing techniques that leverage sequential frames, such as Optical Flow, magnify motion, which facilitates tracking moving objects within a static scene; a different approach is required here, to instead track static objects (UXOs) from a moving perspective (the drone).

## Sequential B-Box Matching and Tubelet Linking

Sequential bounding-box matching is a post-processing method used in object detection to improve the accuracy of predictions, introduced by (Belhassen, et al. 2019). The algorithm matches bounding boxes in adjacent frames by their semantic similarity, and relative locations. The match quality, $q$, is defined by:

$$\frac{1}{q} = \frac{1}{\text{similarity}} = \frac{1}{\text{IoU} \times (Vctr_i \cdot Vctr_j)} \quad (4)$$

where the Intersection over Union (IoU) is multiplied by the dot product of the scoring vectors of the bounding boxes in the denominator. In this implementation, the drone's high velocity combined with a low frame rate frequently resulted in an IoU of 0.0 between frames; so we replaced IoU with the reciprocal of the distance between the boxes' centroids, $1/d_{c1,c2}$. Matched bounding boxes in adjacent frames are candidates for detections of the same object and may form tubelets across multiple frames of the video, as shown in Figure 2. After all the tubelets have been formed, the scoring vectors for the bounding boxes of each tubelet are averaged and used to re-score the classifications. We modified the `detectron2` library to provide scores of all classes per detection, instead of only the highest classification score. This re-scoring successfully leverages the context from multiple frames to improve classification performance.

*Tubelet linking*: After the set of tubelets are created, we can link together two tubelets into a longer tubelet when there is a gap of $n$ frames. The parameter $\kappa$ controls the maximum gap between tubelets to permit linking them. When $\kappa = 1$ this defaults to a gap of size $g = \kappa - 1 = 0$ frames, or just the normal tubelet creation with no linking. With $\kappa = 2$, we link tubelets with a gap of size $g = 1$, and so on.

*Filtering to remove false alarms*: Given the speed that the drone is flying, we know that a UXO will come into the drone's field of view for about 10 frames before exiting the field of view. Perfect tubelets, would therefore typically have a length of about 10 frames. In our application, we found that filtering out short tubelets slightly reduced our false alarm density, but it also adversely affected our peak probability of detection. We decided against applying any tubelet length filtering for this application.

## Performance Results

We analyzed 21 unique runway configurations, containing a total of 410 target UXO opportunities (an average of ~19 UXOs per runway, and each runway had an area of about 4,270 m$^2$). We present our detection performance results in the form of receiver operating characteristics curves that plot our ATR system's probability of target detection as a function of false alarm density, as shown in Figure 3. Each point in the plots corresponds to a different confidence threshold, with higher thresholds on the left.

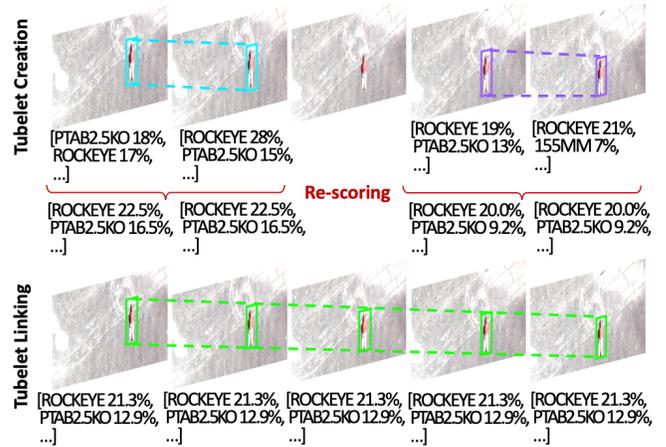

Figure 2: (*Top*) Example of tubelet creation with box re-scoring, and (*Bottom*) tubelet linking with a gap of one frame ($\kappa = 2$) with box re-scoring.

Similarly, we present our classification performance results in the form of a receiver operating characteristics curve that plots the probability of correct classification as a function of false alarm density, as shown in Figure 4. We see that if the model can detect an object, it can classify the detected box consistently well ($P_c > 60\%$). To note, BLU26 is occasionally not detected, which is different from a misclassification. This means that the detection algorithm, stage 1 in CenterNet2, struggles to detect the object prior to the classification task. This is possibly due to the BLU26 sometimes being similar in color to the runway causing edge detection to fail, resulting in no detection and thus no classification.

The confusion matrix, Figure 5, illustrates the performance in terms of correct classification being strong for some classes and less performant in others. Specifically, the 155MM shell struggles compared to others. The 155MM shell is lacking in visible features to describe with text, aside from its overall shape and color. This means that the definitions provided end up also describing other UXOs. For example, it is said in all descriptions that the 155MM is cylindrical or conical and olive green. While there are limited characteristics beyond those that describe the 155MM, those characteristics are not unique to 155MM. The PTAB2.5KO and ROCKEYE specifically are both cylindrical with a nose piece and therefore 155MM are frequently classified as such. Similarly, the PTAB2.5KO, ROCKEYE, and BLU97 are frequently misclassified as each other for the same reason in that they are cylindrical and contain a tail piece.

When testing the MM-OVOD on the novel UXO classes in our dataset with natural language descriptions, it achieved a weighted AP of 0.75, and a weighted average F1-Score of 0.69. These results are particularly impressive as the MM-OVOD model had not been trained on any of the novel UXO classes before.

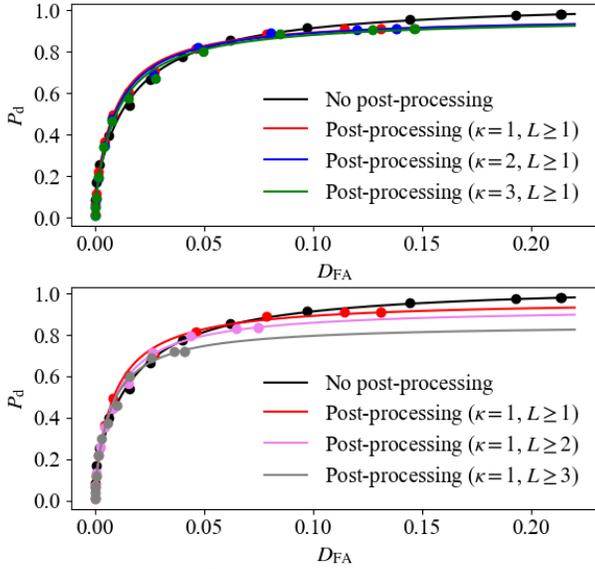

Figure 3: Probability of target detection versus density of false alarms (in m$^2$). *Top*: Tubelet creation and box rescoring improves our results slightly, but tubelet linking did not help in this application. *Bottom*: Filtering out short tubelets removes some false alarms, slightly pushing the curve to the left, but decreases $P_d$.

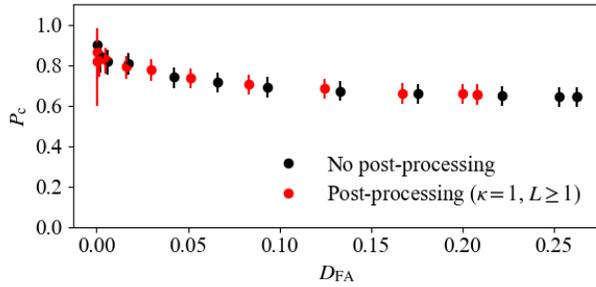

Figure 4: Our ATR system's probability of correct classification averages above 60% for low confidence thresholds and improves as the confidence threshold increases.

Table 2: Performance metrics at a confidence threshold of 0.1, on a test dataset consisting of 21 runways totaling 410 UXO objects (an average of ~19 UXOs per runway).

|  | Precision | Recall | F1-score | Support |
|---|---|---|---|---|
| 155MM | 1.00 | 0.28 | 0.43 | 36 |
| BLU26 | 0.94 | 0.78 | 0.85 | 100 |
| BLU63 | 0.90 | 0.95 | 0.92 | 19 |
| BLU97 | 0.42 | 0.59 | 0.49 | 59 |
| PTAB2.5KO | 0.69 | 0.72 | 0.70 | 95 |
| ROCKEYE | 0.69 | 0.67 | 0.68 | 101 |
| Accuracy |  |  | 0.68 | 410 |
| Macro Avg | 0.66 | 0.57 | 0.58 | 410 |
| Weighted Avg | 0.75 | 0.68 | 0.69 | 410 |

Figure 5: Confusion matrix for conf. threshold = 0.1.

## Mosaic Visualization and Kernel Density Estimation

The sequential bounding-box approach above is limited to sequences of consecutive frames. We developed a novel approach that enables us to consider additional views of the same region that are separated in time (non-contiguous sequences of frames). This occurs when the drone scans the runway using a "lawnmower" pattern and sees an object on one pass, then turns around and sees the same object from the other direction. We expect an increase in both detection and classification performance for cases where an object may be occluded or caught in a shadow from one perspective but clear when viewed from another. Importantly, our false-alarm density, $D_d^{FA}$, can only be computed correctly when multiple passes over the same area are aligned, to not double-count false UXO detections at the same locations (to properly normalize to area). As such, the $D_d^{FA}$ quoted above are *overestimates*, and we will compute the true reduced $D_d^{FA}$ using this mosaic technique in future work.

A classic solution to this problem is to use homographic transformations based off a starting image, such as the solution presented by (Jain and Thapar 2022). This solution works well for images where the downward angle of the camera for the starting image is exactly orthogonal to the surface, i.e., photographing from directly overhead. When the initial frame was taken even just a few degrees from orthogonal, the result is a diminution (or expansion) as the subsequent frames are squeezed (or stretched) to continue the perspective of the first frame. In addition, for the case of object detection from overlapping frames, it is critical that the detected object's locations within the frames are transformed to the same location in the mosaic, rather than just the focusing on the stitched borders. No publicly-available solution has accounted for keeping track of the transformations by frame to be able to transform object locations.

We solved the combination of the diminution and object matching problems by using a two-step procedure in which an initial rough pass is made using a translational transformation, so that a rough mosaic is created without any diminution. We perform a second pass using the output of the first pass as the base image. While performing the stitching, the homographic transformation and translation matrices are saved into a list for each frame. These transformations are then applied to all the detections on each image such that each detection lines up with its location in the mosaic.

Next, we make an overlay by building a 2D Gaussian Kernel Density Estimate with the Gaussians' $\sigma_x$ and $\sigma_y$ proportional to the detected bounding boxes' widths and heights, and Gaussian amplitudes equal to the classification confidences. This overlay aggregates where the ATR System detected targets in subsequent frames *and* multiple drone passes, with the brightest spots representing areas with the highest cumulative confidence from all overlapping frames. The result is a visualization, Figure 6, of the search area that is less sensitive to one-off false alarms and gives the user a holistic overview of the field.

## Path to Deployment

Draper is a non-profit research and development laboratory, and we focus on providing mission-ready technology and capabilities to the U.S. Government. As demonstrated at the ONR UXO test and evaluation event, this ATR software is currently ready to be deployed. If this model were to be deployed on autonomous systems or edge devices, as many of our ATR models at Draper are, some next steps could include structured pruning of the model to remove parameters or weights that have little to no impact on accuracy, or quantization to reduce the precision (storage size) of the weight values (Jin, et al. 2024). Other applications may use sensors in addition to optical imaging (such as LiDAR, SAR, or IR) which could be integrated with the visual language data to perform ATR in situations where direct line of sight is not available (e.g., through cloud cover or in the dark).

We currently have several proposals submitted to DoD customers to deploy our vision-language-based ATR system into a current Government program or platform, and to develop it further. Multiple potential customers have expressed interest in our system's generalizability to new use cases with little (or no) labeled training data and see them as a complementary capability to ATR models trained on specific, large, labeled datasets. In addition to improving the model itself, we are working to improve the user experience for non-technical end-users to add or adjust classes just before runtime. We are also exploring integrating the ATR system into a trusted interface such as Team Awareness Kit (also known as Tactical Assault Kit, or TAK), which is already an integral part of many potential users' workflows.

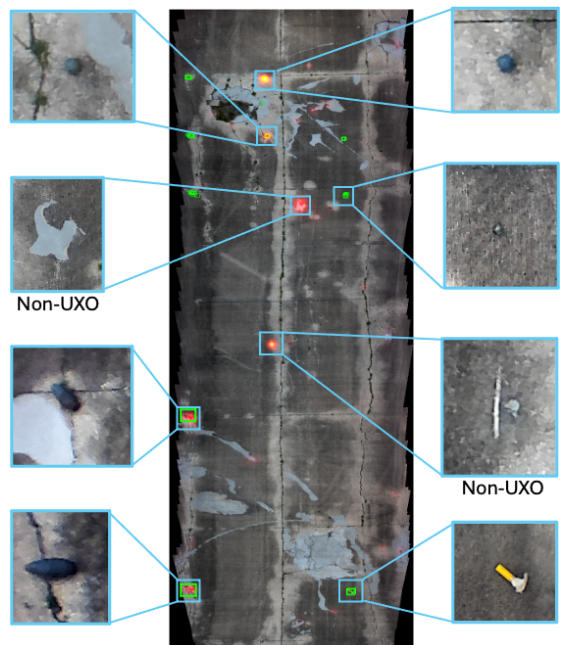

Figure 6: Example mosaic stitched from 100 frames. The heatmap overlay corresponds to aggregated detections across several frames. Green boxes denote ground-truth locations of UXOs.

## Conclusion

ATR systems have many applications and are complicated, costly, and often require bespoke solutions. In this paper, we presented a novel ATR system built around open-vocabulary object detection to enable compatibility with numerous applications. By allowing the end user to define their desired targets of interest "in the field" or just before run time, not only is our system agnostic to the ATR application, it may actually outperform ATR systems based on traditional object detection models for rare or nuanced targets. Further, traditional approaches are brittle, and performance drops as classes evolve (e.g., if the next generation of land-mines are triangular, instead of circular). Our approach allows end-users to quickly add new classes, or adjust existing classes, by simply modifying a few natural language text descriptions, and/or uploading a new exemplar image, thus keeping the technological solution viable longer. We demonstrated good performance, yielding a weighted-average F1-score of 0.69, using a pre-trained VLM and defining our target classes each with just a few natural language descriptions.

## Acknowledgments

We would like to thank ONR for the data, the use case, and the opportunity to participate in the RLAC competition. We also thank Draper for providing the Internal Research & Development funding that supported this work.